\documentclass{article} 
\usepackage{nips13submit_e,times}
\usepackage{hyperref}
\usepackage{url}
\usepackage{graphics}
\usepackage{amsmath,graphicx}
\usepackage{color}
\usepackage{fancyhdr}
\usepackage{algorithm}
\usepackage{algorithmicx}
\usepackage{algpseudocode}
\makeatletter
\renewcommand{\ALG@beginalgorithmic}{\footnotesize}
\makeatother
\usepackage{amssymb}
\usepackage{amsthm}
\graphicspath{{./Figures/}}

\title{Learning Input and Recurrent Weight Matrices in Echo State Networks}

\author{
Hamid Palangi\\
University of British Columbia\\
Vancouver, BC, Canada\\
\texttt{hamidp@ece.ubc.ca} 
\And
Li Deng \\
Microsoft Research \\
Redmond, WA, USA \\
\texttt{deng@microsoft.com} 
\And
Rabab K Ward \\
University of British Columbia\\
Vancouver, BC, Canada\\
\texttt{rababw@ece.ubc.ca} 
}

\begin{document}

\maketitle

\vspace{-20pt}

\begin{abstract}
Echo State Networks (ESNs) are a special type of the temporally deep network model, the Recurrent Neural Network (RNN), where the recurrent matrix is carefully designed and both the recurrent and input matrices are fixed. An ESN uses the linearity of the activation function of the output units to simplify the learning of the output matrix. In this paper, we devise a special technique that take advantage of this linearity in the output units of an ESN, to learn the input and recurrent matrices. This has not been done in earlier ESNs due to their well known difficulty in learning those matrices. Compared to the technique of BackPropagation Through Time (BPTT) in learning general RNNs, our proposed method exploits linearity of activation function in the output units to formulate the relationships amongst the various matrices in an RNN. These relationships results in the gradient of the cost function having an analytical form and being more accurate. This would enable us to compute the gradients instead of obtaining them by recursion as in BPTT. Experimental results on phone state classification show that learning one or both the input and recurrent matrices in an ESN yields superior results compared to traditional ESNs that do not learn these matrices, especially when longer time steps are used.

\end{abstract}

\section{Introduction}
\label{sec:intro}
Recurrent Neural Networks (RNNs) belong to a general type of deep neural networks which are used to model time sequences and dynamical systems \cite{RNNref,robinson1994application,deng1994analysis,mikolov2010recurrent,graves2012sequence,bengio2013advances}. 
Echo State Networks (ESNs) belong to the general class of the RNNs \cite{jaeger2001echostate,jaeger2001short,ESNscience,TriefenbachNIPS2009}.
The following properties of an ESN make it distinct from other types of RNNs. First, both the recurrent and input matrices in an ESN are fixed and not learned (this is largely due to the
difficulty in learning RNN \cite{Bengio_vanish_explode_gradient,pascanu2012difficulty}. Second, the number of hidden neurons in an ESN is typically much larger than those in regular RNNs. The main challenges of training RNNs described in \cite{Bengio_vanish_explode_gradient,pascanu2012difficulty} is avoided in ESN by not training most of the very large number of difficult network parameters. This also leads to avoiding
potential overfitting problems. Third, the output units, also called readout units, in an ESN are linear. This is unlike the typically nonlinear output units in regular RNNs. Given the very large number of hidden neurons of an ESN, the output or readout weight matrix is very large as well. The use of linear output units allows the output weight matrix to be learned very efficiently and with a simple regularization mechanism based on ridge regression. Fourth, the learning of the ESN parameters (i.e. output matrix) is much simpler than that for regular RNNs. The former uses linear learning with convex optimization, and the latter, is based typically on
Back Propagation Through Time (BPTT) and is highly nonlinear and non-convex. As a result, learning the ESN parameters can be effectively carried out via batch training. This greatly facilitates parallel implementation. Learning the general RNN parameters on the other hand, typically requires stochastic gradient descent, and is more difficult for parallelization.

The simplicity in ESN learning comes at the cost of not learning some important parameters (including the input and the recurrent weight matrices) and of using linear output units. 
While the special design of the recurrent matrices (see \cite{RNNtutorial}) and the use of a large number of hidden neurons help in reducing the weakness of using fixed parameters, it is desirable to make these parameters adapt to the data. This is as long as the required learning remains simpler and more parallelizable than the common BPTT learning method applied to regular RNNs. 

This paper presents this type of learning for the input and recurrent matrices of ESNs. We propose a technique that makes full use of the linearity in the output units when constructing constraints on all three input, output, and recurrent matrices in an ESN. The constraints enable us to compute the gradients as the learning signal in an analytical form, and this makes the gradient estimate more
accurate than when computed by recursion as in BPTT. Our preliminary experimental results on phone classification are highly positive.  It is demonstrated that learning one or both of the input and recurrent matrices in an ESN gives better phone classification accuracy than that obtained by a traditional ESN without learning them. Furthermore, when longer time steps are used in analytically computing the gradients, the better classification results are obtained. 

\section{General Recurrent Networks and Specific Echo State Networks}
\label{sec:ESN}

\begin{figure}[h]
\center
   \includegraphics[height=1in,width=5.5in]{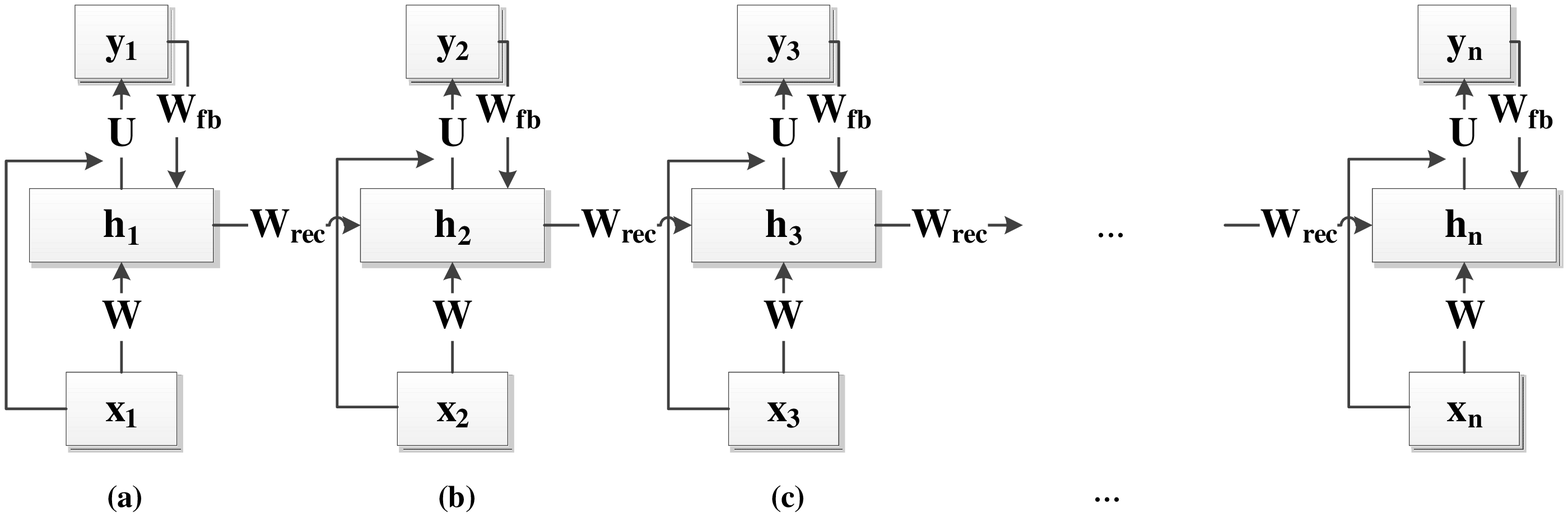}
\vspace{-20pt}
   \caption{Illustration of a general RNN unfolded over time}
   \label{fig:RNNgeneral}
\end{figure}

A general RNN has temporal connections  as well as input-to-hidden layer, hidden layer-to-output connections. These connections are mathematically
represented by the recurrent weight matrix $\mathbf{W}_{{rec}}$, the input weight matrix $\mathbf{W}$, and the output  weight matrix $\mathbf{U}$, respectively. 
The RNN architecture is illustrated in Fig. \ref{fig:RNNgeneral}. It also includes input-to-output and output-to-hidden (feedback) connections, with the latter denoted by $\mathbf{W}_{{fb}}$.
The sequential sections of Fig. \ref{fig:RNNgeneral}(a), \ref{fig:RNNgeneral}(b), \ref{fig:RNNgeneral}(c), \dots, denote the RNN as it unfolds in time. Note that all the weight matrices are constrained to be the same (i.e. they are tied) at any discrete point in time.

In Figure 1, $\mathbf{x}_{i}$, $\mathbf{h}_{i}$ and $\mathbf{y}_{i}$ represent the input, hidden and output vectors at discrete time $t = i$. Again, the connections between the input ($\mathbf{x}_{i}$) and hidden ($\mathbf{h}_{i}$) layers, the hidden  and output ($\mathbf{y}_{i}$) layers, and the output and hidden layers are represented by $\mathbf{W}$, $\mathbf{U}$ and $\mathbf{W}_{{fb}}$ respectively. The temporal connection between $\mathbf{h}_{i}$ and $\mathbf{h}_{{i+1}}$ is represented by the matrix $\mathbf{W}_{{rec}}$. Note that the direct connections from the input to the output layer form a part of the matrix $\mathbf{U}$; i.e., it is equivalent to concatenating the input layer with the hidden layer. 

There are two standard methods to train RNNs, BPTT and the method based on Extended Kalman Filtering (EKF) \cite{RNNtutorial}. BPTT is a first order method, which extends error backpropagation for feed-forward networks by treating each time step as a new hidden layer ( but ties all the weight matrices across time). Generally, BPTT has slow convergence. Its difficulty in capturing long-term memory due to vanishing gradient and exploding gradient problems has been well known for many years \cite{Bengio_vanish_explode_gradient}. It is often non-trivial to obtain good results with BPTT; see the 
tremendous amount of engineering required to make BPTT work \cite{robinson1994application,mikolov2011extensions,sutskever2013training}. The EKF-based method, on the other hand, has fast convergence properties and belongs to a second-order method. However, the computational requirements of the EKF method are high and its implementation is non-trivial \cite{RNNtutorial}, especially for large-scale problems.

One prominent approach that has been proposed to overcome the difficulty in training RNNs is the ESN \cite{ESNscience},\cite{RNNtutorial}. As explained above, an ESN is a special type of RNNs whose recurrent weights ($\mathbf{W}_{{rec}}$) and input to hidden layer weights ($\mathbf{W}$) are fixed and only $\mathbf{U}$ ( that represents the hidden layer to output and the input to output weights ) is trained. The recurrent connections in $\mathbf{W}_{{rec}}$ are sparse and their values are carefully fixed in a way that the echo-state property is preserved. ESNs can be trained very fast because the only connections that are trained are the output connections. With good initialization, ESNs have been shown to yield good performance for one dimensional sequences; but for high dimensional data such as speech, the studies have been relatively limited; please see \cite{TriefenbachNIPS2009}.

Since the output units in an ESN have a linear activation function and assuming the hidden units have a sigmoid activation function, 
the formulation of an ESN as a special type of RNNs ( as shown in Fig. \ref{fig:RNNgeneral} ) can be succinctly described by
\begin{equation}
\label{eq:RNN_h}
\mathbf{h}_{i+1} = \sigma(\mathbf{W}^T\mathbf{x}_{{i+1}}+\mathbf{W}_{rec}{\bf h}_i +\mathbf{W}_{fb}{\bf y}_i)
\end{equation} 
\begin{equation}
\label{eq:RNN_y}
\mathbf{y}_{{i+1}} = \mathbf{U}^T\mathbf{h}_{{i+1}}
\end{equation} 
where $\sigma(x) = \frac{1}{1+e^{-x}}$. As discussed earlier, an ESN has a special designed recurrent matrix, which is endowed with the echo state property in most existing versions \cite{RNNtutorial,ESNscience,TriefenbachNIPS2009}. The echo state property implies that the state, or the hidden units' activities, of the network can be determined uniquely based on the current and previous inputs and outputs provided the network has been running for a sufficiently long time. The formal definition of echo states is described in \cite{RNNtutorial}.
Assume that the maximum eigenvalue of $\mathbf{W}_{{rec}}$ is $\lambda_{max}$, the activation function of the hidden units is sigmoid and $\mid\lambda_{max}\mid > 4$, then the network does not have echo states. This is a sufficient condition for the echo states to not exist\footnote{Note that in \cite{RNNtutorial} the condition is stated as $\mid\lambda_{max}\mid > 1$ because the activation function considered is hyperbolic tangent.} \cite{RNNtutorial}. However, as emphasized in \cite{RNNtutorial}, in practice, when $\mid\lambda_{max}\mid < 4$ the network has echo states. There is a similar sufficient condition under which the exploding gradient problem for recurrent weights would not happen \cite{RazvanRNN}.

In ESN training, only the output weight matrix, $\mathbf{U}$, is trained. The input and  recurrent weight matrices should be carefully fixed. There are three main steps in training an ESN: constructing a network with the echo state property, computing the network states, and estimating the output weights. 

To construct a network with the echo state property, the input weight matrix $\mathbf{W}$ and the sparse recurrent weight matrix $\mathbf{W}_{{rec}}$ are randomly generated. Then, the maximum eigenvalue of $\mathbf{W}_{{rec}}$ is calculated and all entries of $\mathbf{W}_{{rec}}$ are renormalized as follows:
\begin{equation}
\label{eq:WresNormalize}
\mathbf{W}_{{rec}} = \lambda\frac{\mathbf{W}_{{rec}}}{\lambda_{max}}
\end{equation}
where $\lambda < 4$ for sigmoid activation function. $\lambda$ is also an important parameter which affects the memory length of the network and should be determined based on the required memory size for the specific task. As emphasized in \cite{RNNtutorial}, the entries of the input weight matrix are also of great importance; large entries cause most hidden units to saturate while small entries lead hidden units to stay in the linear region of the sigmoid function. The value of $\lambda$ should be predetermined and fixed ( based on the desired task ) before the second step (that calculates the network states).

To find the outputs of the hidden layer units, the hidden states are initialized to zero or another initial state. Then the network runs freely for $i_{trans}$ time steps where the hidden states of each time step are calculated using \eqref{eq:RNN_h} with $\mathbf{W}_{{fb}}=0$. After $i_{trans}$ time steps, the hidden state vectors are stacked in matrix $\mathbf{H}$, i.e.
 \begin{equation}
 \label{eq:Hmatrix}
\mathbf{H} = [\mathbf{h}_{i_{trans}} \mathbf{h}_{i_{trans}+1} \dots \mathbf{h}_{{N}}] 
 \end{equation}
where $N$ is the number of time steps.

To calculate the output weights $\mathbf{U}$, we stack the desired outputs or targets  corresponding to input signal $\mathbf{x}_{i}$ as a matrix $\mathbf{T}$; i.e.,
 \begin{equation}
 \label{eq:Tmatrix}
\mathbf{T} = [\mathbf{t}_{i_{trans}} \mathbf{t}_{i_{trans}+1} \dots \mathbf{t}_{{N}}] 
 \end{equation}

Since $\mathbf{H}$ is computed using all known quantities (incluidng the fixed input and recurrent matrices) using \eqref{eq:RNN_h} and hence it is known also, $\mathbf{U}$ can be obtained by minimizing the following mean-square-error cost function:
 \begin{equation}
 \label{eq:CostFunction}
E = \parallel \mathbf{U}^T\mathbf{H}_{c} - \mathbf{T} \parallel _{F}^{2} = tr[(\mathbf{U}^T\mathbf{H}_{c} - \mathbf{T})(\mathbf{U}^T\mathbf{H}_{c} - \mathbf{T})^T]
 \end{equation}
where $F$ stands for the Frobenius norm of a matrix, $tr(.)$ is the trace of a matrix and 
 \begin{equation}
 \label{eq:Hc}
 \begin{split}
&\mathbf{H}_{c} = [\mathbf{H} \; \mathbf{X}]\\
&\mathbf{X} = [\mathbf{x}_{{i_{trans}}} \mathbf{x}_{{i_{trans}+1}} \dots \mathbf{x}_{{N}}] 
\end{split}
 \end{equation}
Minimizing \eqref{eq:CostFunction}, we have the globally optimal estimate of $\mathbf{U}$ determined by setting the gradient of the above cost function to zero and solving it; i.e.,
 \begin{equation}
 \label{eq:calcU}
 \begin{split}
&\frac{\partial E}{\partial U} = 2\mathbf{H}_{c}(\mathbf{U}^T\mathbf{H}_{c}-\mathbf{T})^T = 0\\
&\mathbf{U} = (\mathbf{H}_{c}\mathbf{H}_{c}^T)^{-1}\mathbf{H}_{c}\mathbf{T}^T
\end{split}
 \end{equation}
In practical implementation, to prevent inaccurate results when $\mathbf{H}\mathbf{H}^T$ is singular or close to singular, the following solution of ``ridge regression'' is used for estimating $\mathbf{U}$:
 \begin{equation}
 \label{eq:calcUwithMu}
\mathbf{U} = (\mathbf{H}_{c}\mathbf{H}_{c}^T + \mu \mathbf{I})^{-1}\mathbf{H}_{c}\mathbf{T}^T
 \end{equation}
where $\mathbf{I}$ is the identity matrix and $\mu$ is a fixed positive number. 
 
\section{Learning the Input Weight Matrix in ESN}
\label{sec:Win}
Assuming the memory of the network extends back to $m$ time steps, we use the following notation to facilitate the development of the 
learning method for the input weight matrix $\mathbf{W}$:
 \begin{equation}
 \label{eq:Notation}
\begin{split}
&\mathbf{X}_{1} = [\mathbf{x}_{1}\;\mathbf{x}_{{m+1}}\;\mathbf{x}_{{2m+1}}\dots],\; \mathbf{X}_{2} = [\mathbf{x}_{2}\;\mathbf{x}_{{m+2}}\;\mathbf{x}_{{2m+2}}\dots], \; \dots \\
&\mathbf{H}_{1} = [\mathbf{h}_{1}\;\mathbf{h}_{{m+1}}\;\mathbf{h}_{{2m+1}}\dots],\; \mathbf{H}_{2} = [\mathbf{h}_{2}\;\mathbf{h}_{{m+2}}\;\mathbf{h}_{{2m+2}}\dots], \; \dots \\
&\mathbf{T}_{1} = [\mathbf{t}_{1}\;\mathbf{t}_{{m+1}}\;\mathbf{t}_{{2m+1}}\dots],\; \mathbf{T}_{2} = [\mathbf{t}_{2}\;\mathbf{t}_{{m+2}}\;\mathbf{t}_{{2m+2}}\dots], \; \dots 
\end{split}
 \end{equation}
Therefore, equations \eqref{eq:RNN_h} and \eqref{eq:RNN_y} can be written as:
\begin{equation}
\label{eq:RNN_Hm}
\mathbf{H}_{{i+1}} = \sigma(\mathbf{W}^T\mathbf{X}_{{i+1}}+\mathbf{W}_{rec}{\bf H}_i)
\end{equation} 
\begin{equation}
\label{eq:RNN_Ym}
\mathbf{Y}_{{i+1}} = \mathbf{U}^T\mathbf{H}_{{i+1}}
\end{equation} 

To find the gradient of the cost function $E$ with respect to $\mathbf{W}$ and learn input weights $\mathbf{W}$ we consider two cases in the remainder of this section. In Case 1, we assume that $\mathbf{U}$ does not depend on $\mathbf{W}$ and in Case 2 we take into account the dependency between $\mathbf{U}$ and $\mathbf{W}$. Note that in both cases we take into account the time dependency among the hidden state vectors in $\mathbf{H}$, i.e., the dependency of $\mathbf{h}_{{i+1}}$ on $\mathbf{h}_{i}$ at every time step $i$ \footnote{Note that this is one of the main differences with the work presented in \cite{DengYu2011,SingleLayer} where there is no temporal connection in the single layer network and hence no time dependency is considered.}. Since Case 2 is a more realistic formulation of the gradient, it is used for learning the input weight matrix  in our experimental results presented in Section 5. We derive the gradient for one time step dependency and then generalize it to $n$ time step dependency, i.e., $\mathbf{H}_{i}$ depends on $\mathbf{H}_{{i-1}}$, $\mathbf{H}_{{i-2}}$ and so on up to $n$ time steps.

\subsection{Case 1}
\label{subsec:case1}
The gradient of the cost function with respect to $\mathbf{W}$ can be written as 
 \begin{equation}
 \label{eq:case1eq1}
 \begin{split}
&\frac{\partial E} {\partial \mathbf{W}} = \frac{\partial} {\partial \mathbf{W}}tr[(\mathbf{U}^T\mathbf{H}_{2}-\mathbf{T}_{2})(\mathbf{U}^T\mathbf{H}_{2}-\mathbf{T}_{2})^T]\\
&= \frac{\partial} {\partial \mathbf{W}}tr[(\mathbf{U}^T\sigma(\mathbf{W}_{rec}\mathbf{H}_{1}+\mathbf{W}^T\mathbf{X}_{2})-\mathbf{T}_{2})(\mathbf{U}^T\sigma(\mathbf{W}_{rec}\mathbf{H}_1+\mathbf{W}^T\mathbf{X}_2)-\mathbf{T}_2)^T]\\
&= [\frac{\partial} {\partial \mathbf{W}}\sigma(\mathbf{W}_{rec}\mathbf{H}_1+\mathbf{W}^T\mathbf{X}_{2})][2\mathbf{U}^T(\mathbf{U}^T\mathbf{H}_{2}-\mathbf{T}_{2})^T]
\end{split}
 \end{equation}
Assuming that $\mathbf{H}_{1}$ depends on $\mathbf{W}$, i.e., $\mathbf{H}_{1}=\sigma(\mathbf{W}_{rec}\mathbf{H}_{0}+\mathbf{W}^T\mathbf{X}_{1})$, and denoting the term independent of $\mathbf{W}$ by:
 \begin{equation}
 \label{eq:case1eq2}
\mathbf{S} = 2\mathbf{U}^T(\mathbf{U}^T\mathbf{H}_{2}-\mathbf{T}_{2})^T
 \end{equation}
then using  chain rule of calculus  we have:
 \begin{equation}
 \label{eq:case1eq3}
\frac{\partial E} {\partial \mathbf{W}} = [\frac{\partial} {\partial \mathbf{W}}[\mathbf{W}_{{rec}}\sigma(\mathbf{W}_{rec}\mathbf{H}_{0}+\mathbf{W}^T\mathbf{X}_{1})+\mathbf{W}^T\mathbf{X}_{2}]]\mathbf{H}_{2}^T\circ(\mathbf{1}-\mathbf{H}_{2}^T)\circ\mathbf{S}
 \end{equation}
and therefore
 \begin{equation}
 \label{eq:case1eq4}
\frac{\partial E} {\partial \mathbf{W}} = \mathbf{X}_{1}[\mathbf{H}_{1}^T\circ(\mathbf{1}-\mathbf{H}_{1}^T)\mathbf{W}_{{rec}}^T\circ\mathbf{H}_{2}^T\circ(\mathbf{1}-\mathbf{H}_{2}^T)\circ\mathbf{S}] + \mathbf{X}_{2}[\mathbf{H}_{2}^T\circ(\mathbf{1}-\mathbf{H}_{2}^T)\circ\mathbf{S}]
 \end{equation}
 where $\circ$ is element-wise multiplication.
\subsection{Case 2}
\label{subsec:case2}
The gradient calculated in Case 1 is not accurate because the dependency between $\mathbf{U}$ and $\mathbf{W}$ is ignored. To take this dependency into consideration, the first line of equation \eqref{eq:case1eq1} is rewritten as:
 \begin{equation}
 \label{eq:case2eq1}
\frac{\partial E} {\partial \mathbf{W}} = \frac{\partial} {\partial \mathbf{W}}tr(\underbrace{\mathbf{U}^T\mathbf{H}_{2}\mathbf{H}_{2}^T\mathbf{U}}_{a}-\underbrace{\mathbf{U}^T\mathbf{H}_{2}\mathbf{T}_{2}^T}_{b}-\mathbf{T}_{2}\mathbf{H}_{2}^T\mathbf{U}+\mathbf{T}_{2}\mathbf{T}_{2}^T)
 \end{equation}
Substituting $\mathbf{U} = (\mathbf{H}_{2}\mathbf{H}_{2}^T)^{-1}\mathbf{H}_{2}\mathbf{T}_{2}^T$ we have
 \begin{equation}
 \label{eq:case2eq2}
 \begin{split}
& a = \mathbf{T}_{2}\mathbf{H}_{2}^T(\mathbf{H}_{2}\mathbf{H}_{2}^T)^{-T}\underbrace{\mathbf{H}_{2}\mathbf{H}_{2}^T(\mathbf{H}_{2}\mathbf{H}_{2}^T)^{-1}}_{\mathbf{I}}\mathbf{H}_{2}\mathbf{T}_{2}^T\\
& = \mathbf{T}_{2}\mathbf{H}_{2}^T(\mathbf{H}_{2}\mathbf{H}_{2}^T)^{-T} \mathbf{H}_{2}\mathbf{T}_{2}^T
\end{split}
 \end{equation}
 and
 \begin{equation}
 \label{eq:case2eq3}
b = \mathbf{T}_{2}\mathbf{H}_{2}^T(\mathbf{H}_{2}\mathbf{H}_{2}^T)^{-T} \mathbf{H}_{2}\mathbf{T}_{2}^T
 \end{equation}
 therefore
 \begin{equation}
 \label{eq:case2eq4}
\frac{\partial E}{\partial \mathbf{W}}=-\frac{\partial}{\partial \mathbf{W}}tr(\mathbf{T}_{2}\mathbf{H}_{2}^T(\mathbf{H}_{2}\mathbf{H}_{2}^T)^{-1}\mathbf{H}_{2}\mathbf{T}_{2}^T)
 \end{equation}
Since $tr(\mathbf{AB})=tr(\mathbf{BA})$ and $tr(\mathbf{A}) = tr(\mathbf{A}^T)$ we have
 \begin{equation}
 \label{eq:case2eq5}
\frac{\partial E}{\partial \mathbf{W}}=-\frac{\partial}{\partial \mathbf{W}}tr((\mathbf{H}_{2}\mathbf{H}_{2}^T)^{-1}\mathbf{H}_{2}\mathbf{T}_{2}^T\mathbf{T}_{2}\mathbf{H}_{2}^T)
 \end{equation}
Using the chain rule presented in equation (126) of \cite{MatrixCookBook}, the gradient can be written as:
 \begin{equation}
 \label{eq:case2eq6}
\frac{\partial E}{\partial \mathbf{W}} = -[\underbrace{\frac{\partial}{\partial \mathbf{W}}\mathbf{H}_{2}}_{\mathbf{B}}]
[\underbrace{\frac{\partial}{\partial \mathbf{H}_{2}^T}tr((\mathbf{H}_{2}\mathbf{H}_{2}^T)^{-1}\mathbf{H}_{2}\mathbf{T}_{2}^T\mathbf{T}_{2}\mathbf{H}_{2}^T)}_{\mathbf{A}}]
 \end{equation}
Below we first calculate matrix $\mathbf{A}$ and then matrix $\mathbf{B}$.

For constant values of hidden states $\mathbf{H}_{2}$ we define $\mathbf{F}$, $\mathbf{M}$ and $\mathbf{S}$ as follows:
 \begin{equation}
 \label{eq:case2eq7temp1}
 \begin{split}
&\mathbf{F} = (\mathbf{H}_{2}\mathbf{H}_{2}^T)^{-1}\\
&\mathbf{M}=\mathbf{T}_{2}^T\mathbf{T}_{2}\\
&\mathbf{S} = \mathbf{H}_{2}\mathbf{M}\mathbf{H}_{2}^T
\end{split}
 \end{equation}
Then, using the chain rule again, we obtain
 \begin{equation}
 \label{eq:case2eq7}
\mathbf{A} = 
\underbrace{\frac{\partial}{\partial \mathbf{H}_{2}^T}tr(\mathbf{F}\mathbf{H}_{2}\mathbf{M}\mathbf{H}_{2}^T)}_{\mathbf{A}_{1}} + 
\underbrace{\frac{\partial}{\partial \mathbf{H}_{2}^T}tr((\mathbf{H}_{2}\mathbf{H}_{2}^T)^{-1}\mathbf{S})}_{\mathbf{A}_{2}}
 \end{equation}
considering the fact that $tr(\mathbf{A}\mathbf{B}) = tr(\mathbf{B}\mathbf{A})$, $\mathbf{A_1}$ can be written as
 \begin{equation}
 \label{eq:case2eq8}
\mathbf{A}_{1} = 
\frac{\partial}{\partial \mathbf{H}_{2}^T}tr(\mathbf{M}\mathbf{H}_{2}^T\mathbf{F}\mathbf{H}_{2}) 
 \end{equation}
From equation (107) of \cite{MatrixCookBook} we have
 \begin{equation}
 \label{eq:case2eq9}
\mathbf{A}_{1} = \mathbf{M}^T\mathbf{H}_{2}^T\mathbf{F}^T + \mathbf{M}\mathbf{H}_{2}^T\mathbf{F}
 \end{equation}
Since $\mathbf{F}^T = \mathbf{F}$ and $\mathbf{M}^T = \mathbf{M}$, $\mathbf{A}_{1}$ can be written as
 \begin{equation}
 \label{eq:case2eq10}
\mathbf{A}_{1} = 2\mathbf{M}\mathbf{H}_{2}^T\mathbf{F}
 \end{equation}
Considering equation (114) of \cite{MatrixCookBook}, $\mathbf{A}_{2}$ will be as follows
 \begin{equation}
 \label{eq:case2eq11}
\mathbf{A}_{2} = -(\mathbf{H}_{2}^T(\mathbf{H}_{2}\mathbf{H}_{2}^T)^{-1})(\mathbf{S}+\mathbf{S}^T)(\mathbf{H}_{2}\mathbf{H}_{2}^T)^{-1}
 \end{equation}
substituting $\mathbf{S}$ and using $\mathbf{M} = \mathbf{M}^T$ results in
 \begin{equation}
 \label{eq:case2eq12}
\mathbf{A}_{2} = -2\mathbf{H}_{2}^T(\mathbf{H}_{2}\mathbf{H}_{2}^T)^{-1}(\mathbf{H}_{2}\mathbf{M}\mathbf{H}_{2}^T)(\mathbf{H}_{2}\mathbf{H}_{2}^T)^{-1}
 \end{equation}
Now substituting $\mathbf{M}$ and $\mathbf{F}$ in \eqref{eq:case2eq7temp1} results in the final formulation for $\mathbf{A}$ as follows
 \begin{equation}
 \label{eq:case2eq13}
\mathbf{A} = 2\mathbf{T}_{2}^T\mathbf{T}_{2}\mathbf{H}_{2}^T(\mathbf{H}_{2}\mathbf{H}_{2}^T)^{-1} - 
2\mathbf{H}_{2}^T(\mathbf{H}_{2}\mathbf{H}_{2}^T)^{-1}\mathbf{H}_{2}\mathbf{T}_{2}^T\mathbf{T}_{2}\mathbf{H}_{2}^T(\mathbf{H}_{2}\mathbf{H}_{2}^T)^{-1}
 \end{equation}
 
To calculate $\mathbf{B}$, we assume that there is just one time step dependency, i.e., $\mathbf{H}_{2}$ depends on $\mathbf{H}_{1}$ and $\mathbf{W}$ and $\mathbf{H}_{1}$ does not depend on $\mathbf{H}_{0}$ but depends on $\mathbf{W}$. The generalization to an arbitrary number of time steps is straightforward and is presented at the end of this section. From \eqref{eq:case2eq6} and \eqref{eq:RNN_Hm} we write $\mathbf{B}$ as: 
 \begin{equation}
 \label{eq:case2eq14}
\mathbf{B} = \frac{\partial}{\partial \mathbf{W}}[\sigma( \mathbf{W}_{{rec}}\sigma( \mathbf{W}_{rec} {\bf H}_0 + \mathbf{W}^T\mathbf{X}_{1} ) + \mathbf{W}^T\mathbf{X}_{2} )]
 \end{equation}
which is similar to the term calculated in \eqref{eq:case1eq4} and therefore
 \begin{equation}
 \label{eq:case2eq15}
\mathbf{B} = \mathbf{X}_{1}[\mathbf{H}_{1}^T\circ(\mathbf{1}-\mathbf{H}_{1}^T)\mathbf{W}_{{rec}}^T\circ\mathbf{H}_{2}^T\circ(\mathbf{1}-\mathbf{H}_{2}^T)] + \mathbf{X}_{2}[\mathbf{H}_{2}^T\circ(\mathbf{1}-\mathbf{H}_{2}^T)]
 \end{equation}
By substituting \eqref{eq:case2eq13} and \eqref{eq:case2eq15} in \eqref{eq:case2eq6} we get the gradient formulation for one time step dependency as follows:
 \begin{equation}
 \label{eq:case2eq16}
\frac{\partial E}{\partial \mathbf{W}} = -[\mathbf{X}_{1}[\mathbf{H}_{1}^T\circ(\mathbf{1}-\mathbf{H}_{1}^T)\mathbf{W}_{{rec}}^T\circ\mathbf{H}_{2}^T\circ(\mathbf{1}-\mathbf{H}_{2}^T)\circ \mathbf{A}] + \mathbf{X}_{2}[\mathbf{H}_{2}^T\circ(\mathbf{1}-\mathbf{H}_{2}^T)\circ \mathbf{A}]]
 \end{equation}

This gradient formulation can be generalized for an arbitrary number of time steps as follows: 
 \begin{equation}
 \label{eq:case2eq17}
\frac{\partial E}{\partial \mathbf{W}} = -[\sum\limits_{i=1}^{n}\mathbf{X}_{i}\mathbf{C}_{i}]
 \end{equation}
 where $n$ is the number of time steps and 
 \begin{equation}
 \label{eq:case2eq18}
\begin{split}
&\mathbf{C}_{i} = [\mathbf{H}_{i}^T\circ(\mathbf{1} - \mathbf{H}_{i}^T)\mathbf{W}_{{rec}}^T]\circ \mathbf{C}_{{i+1}} \; \; \; \; , \; for \; \; i = 1 , \dots , n-1\\
&\mathbf{C}_{n} = \mathbf{H}_{n}^T\circ(\mathbf{1} - \mathbf{H}_{n}^T)\circ \mathbf{A}
\end{split}
 \end{equation}
 To calculate $\mathbf{A}$ using \eqref{eq:case2eq13}, $\mathbf{H}_{n}$ and $\mathbf{T}_{n}$ are used. 

After calculating the gradient of the cost function w.r.t $\mathbf{W}$, the input weights $\mathbf{W}$ are updated using the following update equation 
 \begin{equation}
 \label{eq:Wupdate}
\mathbf{W}_{{i+1}} = \mathbf{W}_{i} - \alpha \frac{\partial E}{\partial \mathbf{W}_{i}} + \beta (\mathbf{W}_{i} - \mathbf{W}_{{i-1}} )
 \end{equation}
where $\alpha$ is the step size and
 \begin{equation}
 \label{eq:Beta}
 \begin{split}
&\beta = \frac{m_{old}}{m_{new}}\\
&m_{new} = \frac{1 + \sqrt{ 1 + 4m_{old}^2 }}{2}
\end{split}
 \end{equation}
and where the initial value for $m_{old}$ and $m_{new}$ is $1$. The third term in \eqref{eq:Wupdate} helps the algorithm to converge faster and is based on the FISTA algorithm proposed in \cite{FISTA} and used in \cite{SingleLayer}. 
 
\section{Learning the Recurrent Weight Matrix ($\mathbf{W}_{{rec}}$) in the ESN}
\label{sec:Wres}
To learn the recurrent weights, the gradient of the cost function w.r.t $\mathbf{W}_{{rec}}$ should be calculated. We first derive the formulation for the two time steps dependency, i.e., $\mathbf{H}_{2}$ depends on $\mathbf{H}_{1}$ and $\mathbf{H}_{1}$ depends on $\mathbf{H}_{0}$ but no more time dependencies. Then it will be generalized to the arbitrary number of time steps. The same method that is used in section \ref{sec:Win} can be used to get the following formulation for the gradient:
 \begin{equation}
 \label{eq:WresEq1}
\frac{\partial E}{\partial \mathbf{W}_{{rec}}} = -[\underbrace{\frac{\partial}{\partial \mathbf{W}_{{rec}}}\mathbf{H}_{2}}_{\mathbf{B}}]
[\underbrace{\frac{\partial}{\partial \mathbf{H}_{2}^T}tr((\mathbf{H}_{2}\mathbf{H}_{2}^T)^{-1}\mathbf{H}_{2}\mathbf{T}_{2}^T\mathbf{T}_{2}\mathbf{H}_{2}^T)}_{\mathbf{A}}]
 \end{equation}
$\mathbf{A}$ will be the same as \eqref{eq:case2eq13}. To calculate $\mathbf{B}$ we have
 \begin{equation}
 \label{eq:WresEq2}
 \begin{split}
&\mathbf{B} = \frac{\partial}{\partial \mathbf{W}_{{rec}}} [ \sigma ( \mathbf{W}_{{rec}}\sigma ( \mathbf{W}_{rec}\mathbf{H}_{0} + \mathbf{W}^T\mathbf{X}_{1} ) + \mathbf{W}^T\mathbf{X}_{2} ) ]\\
& = [\frac{\partial}{\partial \mathbf{W}_{{rec}}} [ \mathbf{W}_{{rec}}\underbrace{\sigma ( \mathbf{W}_{rec}\mathbf{H}_0 + \mathbf{W}^T\mathbf{X}_{1} )}_{\mathbf{H}_{1}} + \mathbf{W}^T\mathbf{X}_{2} ]]\mathbf{H}_{2}^T \circ ( \mathbf{1} - \mathbf{H}_{2}^T )\\
& = [ \mathbf{H}_{1} + \mathbf{W}_{{rec}}\underbrace{\mathbf{H}_{0}[ \mathbf{H}_{1}^T \circ ( \mathbf{1} - \mathbf{H}_{1}^T ) ]}_{\frac{\partial \mathbf{H}_{1}}{\partial \mathbf{W}_{{rec}}}} ]\mathbf{H}_{2}^T \circ ( \mathbf{1} - \mathbf{H}_{2}^T )
\end{split}
 \end{equation}
and therefore the gradient will be:
 \begin{equation}
 \label{eq:WresEq3}
\frac{\partial E}{\partial \mathbf{W}_{{rec}}} = \mathbf{H}_{1} [ \mathbf{H}_{2}^T \circ ( \mathbf{1} - \mathbf{H}_{2}^T ) \circ \mathbf{A} ] + \mathbf{W}_{{rec}}\mathbf{H}_{0}[ \mathbf{H}_{1}^T \circ ( \mathbf{1} - \mathbf{H}_{1}^T ) \circ (\mathbf{H}_{2}^T \circ ( \mathbf{1} - \mathbf{H}_{2}^T ) \circ \mathbf{A}) ]
 \end{equation}
It can be generalized for arbitrary number of time steps as follows:
 \begin{equation}
 \label{eq:WresEq4}
\frac{\partial E}{\partial \mathbf{W}_{{rec}}} = \sum\limits_{i=1}^{n} \mathbf{W}_{{rec}}^{n-i}\mathbf{H}_{{i-1}}\mathbf{C}_{i}
 \end{equation}
 where $\mathbf{H}_{0}$ includes the initial hidden states and
 \begin{equation}
 \label{eq:WresEq5}
 \begin{split}
&\mathbf{C}_{n} = \mathbf{H}_{n}^T \circ ( \mathbf{1} - \mathbf{H}_{n}^T ) \circ \mathbf{A}\\
&\mathbf{C}_{i} = \mathbf{H}_{i}^T \circ ( \mathbf{1} - \mathbf{H}_{i}^T ) \circ \mathbf{C}_{{i+1}}
\end{split}
 \end{equation}
 and $\mathbf{A}$ is calculated using \eqref{eq:case2eq13} based on $\mathbf{H}_n$ and $\mathbf{T}_n$.

Only the non-zero entries of the sparse matrix $\mathbf{W}_{{rec}}$ are updated using \eqref{eq:Wupdate} and the gradient calculated in \eqref{eq:WresEq4}. To make sure that the network has the echo state property after each epoch, the entries of $\mathbf{W}_{{rec}}$ are renormalized such that the maximum eigenvalue of $\mathbf{W}_{{rec}}$ is $\lambda$ that is predetermined in \eqref{eq:WresNormalize}. This renormalization also prevents the gradient explosion problem for recurrent weights from happening.
 
A summary of the learning method is as follows:
\begin{itemize}
\item The echo state network with predetermined maximum eigenvalue of $\mathbf{W}_{{rec}}$ is constructed based on the explanations presented in section \ref{sec:ESN}.
\item Input weights matrix $\mathbf{W}$ is updated based on \eqref{eq:case2eq17}, \eqref{eq:case2eq18}, \eqref{eq:Wupdate} and \eqref{eq:Beta}.
\item Non-zero entries of the sparse recurrent weights matrix $\mathbf{W}_{{rec}}$ are updated based on \eqref{eq:WresEq4}, \eqref{eq:WresEq5}, \eqref{eq:Wupdate} for $\mathbf{W}_{{rec}}$ and \eqref{eq:Beta}.
\item Updated $\mathbf{W}_{{rec}}$ is renormalized to have the predetermined maximum eigenvalue $\lambda$.
\item The forward pass is repeated with the updated input and recurrent weights to find the hidden states. The network runs freely for $i_{trans}$ time steps and then the hidden states are recorded as matrix $\mathbf{H}$. 
\item Taking into account the direct connections from the input to output in the network, the output weight matrix $\mathbf{U}$ is calculated using \eqref{eq:calcUwithMu}.
\end{itemize}
To prevent the value of the gradient w.r.t $\mathbf{W}$ from exploding, we use a similar approach proposed in \cite{RazvanRNN} where the gradient value is renormalized when it is greater than a threshold.
 
\section{Experiments}
\label{sec:experiments}
We carried out the experiments for frame-level classification of phone states on the TIMIT dataset using an ESN with all parameters learned as discussed so far. 
The training data includes 1,124,589 frames. The validation set has 122,488 frames from 50 speakers. The results are reported using the core test set consisting of 192 sentences and 57,920 frames. The speech is analysed using the standard Mel Frequency Cepstral Coefficients (MFCC). Each feature vector has 39 entries, including first and second derivatives.
We have used 3 states for each of 61 phones resulting in a target class vector with 183 entries. Phone state labels are extracted using a GMM-HMM system which aligns the frames with their corresponding states. 

We have used a context window of 3 frames for all experiments resulting in the input vectors with $3 \times 39 = 117$ entries. The regularization parameter $\mu$ used in \eqref{eq:calcUwithMu} is set to $10^{-8}$. The maximum eigenvalue of $\mathbf{W}_{{rec}}$ is set to $3.9$. We have used a step size ($\alpha$ in \eqref{eq:Wupdate}) of $0.07$. 
The task is to classify each frame in the TIMIT core test set into one of 183 phone states. 
The results are presented in Table \ref{table:Results} for different hidden layer sizes in the ESN, one for each row in the table.
The results are also arranged by four different ways of learning the input and recurrent weigh matrices $\mathbf{W}$ and $\mathbf{W}_{{rec}}$, where  $m$ is the number of time steps in incorporating matrices' dependencies in the learning. The column of ``ESN'' refers to the traditional ESN as in \cite{jaeger2001echostate,jaeger2001short,ESNscience,TriefenbachNIPS2009} where  $\mathbf{W}$ and $\mathbf{W}_{{rec}}$ are not learned.

\begin{table}[h]
\caption{Frame-level phone-state classification error rates for the TIMIT core test set}
\label{table:Results}
\begin{center}
\begin{tabular}{ | l | l | c |c |c | } 
\hline
    Hidden units & ESN & Learning $\mathbf{W}$ & Learning $\mathbf{W}$ and $\mathbf{W}_{{rec}}$ & Learning $\mathbf{W}$ and $\mathbf{W}_{{rec}}$ \\ 
       &    &  with $m=1$ & with $m=1$ &  with $m=3$ \\ \hline
    100 & 75.5\% & 66.7\% &  64.0 \% & 63.2\%  \\ \hline
	500 & 70.1\% &  59.8\%& 57.5\%  & 56.8\% \\ \hline
	2000 & 63.8\% & 54.2\% & 52.7\% & 52.1\% \\ \hline
	10000 & 57.1\% & 49.5\% & 48.0\% &  46.8\%\\ \hline
	30000 & 53.3\% & 45.9\% & 44.5\%  & 43.0\% \\ \hline		
\end{tabular}
\end{center}
\end{table}

The preliminary experimental results shown in Table verify that learning input and recurrent weight matrices in the ESN is superior to the ESN with the same structure 
but without learning the two matrices. Furthermore, the longer the time steps are incorporated in the learning, the lower error rates are. On the column of ``ESN'', we also observe that
the traditional ESN improves its performance as the number of hidden units increases, consistent with the findings reported in \cite{TriefenbachNIPS2009}. Finally, we would like to remark that the results obtained so far are very preliminary, and the task is on frame-level classification of 183 phone states. Our first step of research is focused on this easiest task since it is a pure and simple machine learning problem and it requires no expertise in speech recognition. The next steps are to move 1)  from 183-state classification to 39-phone classification; 2) from frame level to segment level (which requires dynamic programming over three states of each phone); 
and 3) from classification (with no phone insertion and deletion errors) to recognition (with phone insertion and  deletion errors). Then we can meaningfully compare the results with other approaches in the literature on the TIMIT phone recognition task.

\section{Discussion and Conclusion}
\label{sec:conclusions}

The main idea of this paper is straightforward: the traditional ESN learns only one of three important sets of weight matrices, and we want to learn them all.
The key property that characterizes the ESN is the use of linear output units so that the learning is simple, convex, forms a least-square ridge regression problem with a global optimum (in learning the output weights). In extending the learning of the output weights to only learning input and recurrent weights, we make use of the same property of linear output units to develop and formulate constraints among various sets of ESN weight matrices. Such constraints are then used to derive analytic forms of the error gradients w.r.t the input and recurrent weights to be learned. The standard learning method of BPTT for the general RNN (with typically nonlinear output units) does not admit analytical forms of gradient computation. BPTT requires recursively propagating the error signal backward through time, a very different style of computation and learning than what we have developed in this work for ESNs.

In this paper we focus on ESNs with one layer and with no feedback connection, i.e., $\mathbf{W}_{fb}=0$, without loss of generality. As our future work, to build more layers of ESN, we can simply stack the output of one-layer ESN on top of another, or we can combine the output with the original data input and/or with hidden units. Also, in the current work with one-layer ESN, we take into account the dependency between $\mathbf{U}$ and $\mathbf{W}$, $\mathbf{h}_{i}$ and $\mathbf{W}$, $\mathbf{h}_{i}$ and $\mathbf{h}_{{i-1}}$, etc. When more layers of ESN are built, a richer dependency becomes available to exploit but the same principle used in this work in deriving the analytic forms of the gradient computation would apply to the multiple-layer ESN in our future work.

\newpage
\bibliographystyle{IEEEbib}
\bibliography{refs}

\end{document}